\documentclass[letterpaper, 10 pt, journal, twoside]{IEEEtran}  

\IEEEoverridecommandlockouts                              

\usepackage{amsthm,amsmath,amssymb}
\usepackage{mathrsfs}
\usepackage{graphicx}
\usepackage{float}
\usepackage{cite}
\usepackage{multirow}
\usepackage{diagbox}
\usepackage{stfloats}
\usepackage{multicol}
\usepackage{algorithmic}
\usepackage{algorithm}
\usepackage{inputenc}
\usepackage{subfigure}
\usepackage{bm}
\usepackage{booktabs}

\usepackage[switch]{lineno}  
\usepackage{changepage}

\usepackage{tabularx}

\usepackage[labelsep=period]{caption}
\title{Time-Efficient Mars Exploration of Simultaneous Coverage and Charging with Multiple Drones}

\author{
	Yuan Chang, Chao Yan, Xingyu Liu, Xiangke Wang, Han Zhou, Xiaojia Xiang, Dengqing Tang$^*$
\thanks{The authors are with National University of Defense Technology, Changsha, China. Yuan Chang and Chao Yan contributed equally to this work.}%
\thanks{$^*$ Corresponding author.}
}

\begin{document}


\maketitle

\captionsetup{font={small}}

\begin{abstract}
This paper presents a time-efficient scheme for Mars exploration by the cooperation of multiple drones and a rover. To maximize effective coverage of the Mars surface in the long run, a comprehensive framework has been developed with joint consideration for limited energy, sensor model, communication range and safety radius, which we call TIME-SC$^2$ (TIme-efficient Mars Exploration of Simultaneous Coverage and Charging). First, we propose a multi-drone coverage control algorithm by leveraging emerging deep reinforcement learning and design a novel information map to represent dynamic system states. Second, we propose a near-optimal charging scheduling algorithm to navigate each drone to an individual charging slot, and we have proven that there always exists feasible solutions. The attractiveness of this framework not only resides on its ability to maximize exploration efficiency, but also on its high autonomy that has greatly reduced the non-exploring time. Extensive simulations have been conducted to demonstrate the remarkable performance of TIME-SC$^2$ in terms of time-efficiency, adaptivity and flexibility.

\end{abstract}

\section{Introduction}

Mars exploration has infinite benefits in terms of innovation, culture, and technology\cite{Roadmap}. However, the slow-moving speed and limited sensing range of existing rovers make them inefficient for Mars exploration. For example, Spirit, the most successful rover to date, traveled only 7.73 kilometers in 6 years\cite{Spirit}. There is a clear need to redesign the current rover system for higher exploration efficiency, i.e., to cover more areas in a given time. Meanwhile, the agility and aerial reach of drones make them highly promising for Mars exploration \cite{mars_2020,TERRA,ARCHES}. Recently, NASA launched the rover Perseverance, and unprecedentedly equipped with a drone called Ingenuity, opening a new era for Mars exploration \cite{mars_2020}.

\begin{figure}
	\centering
	\includegraphics[width=3.3in]{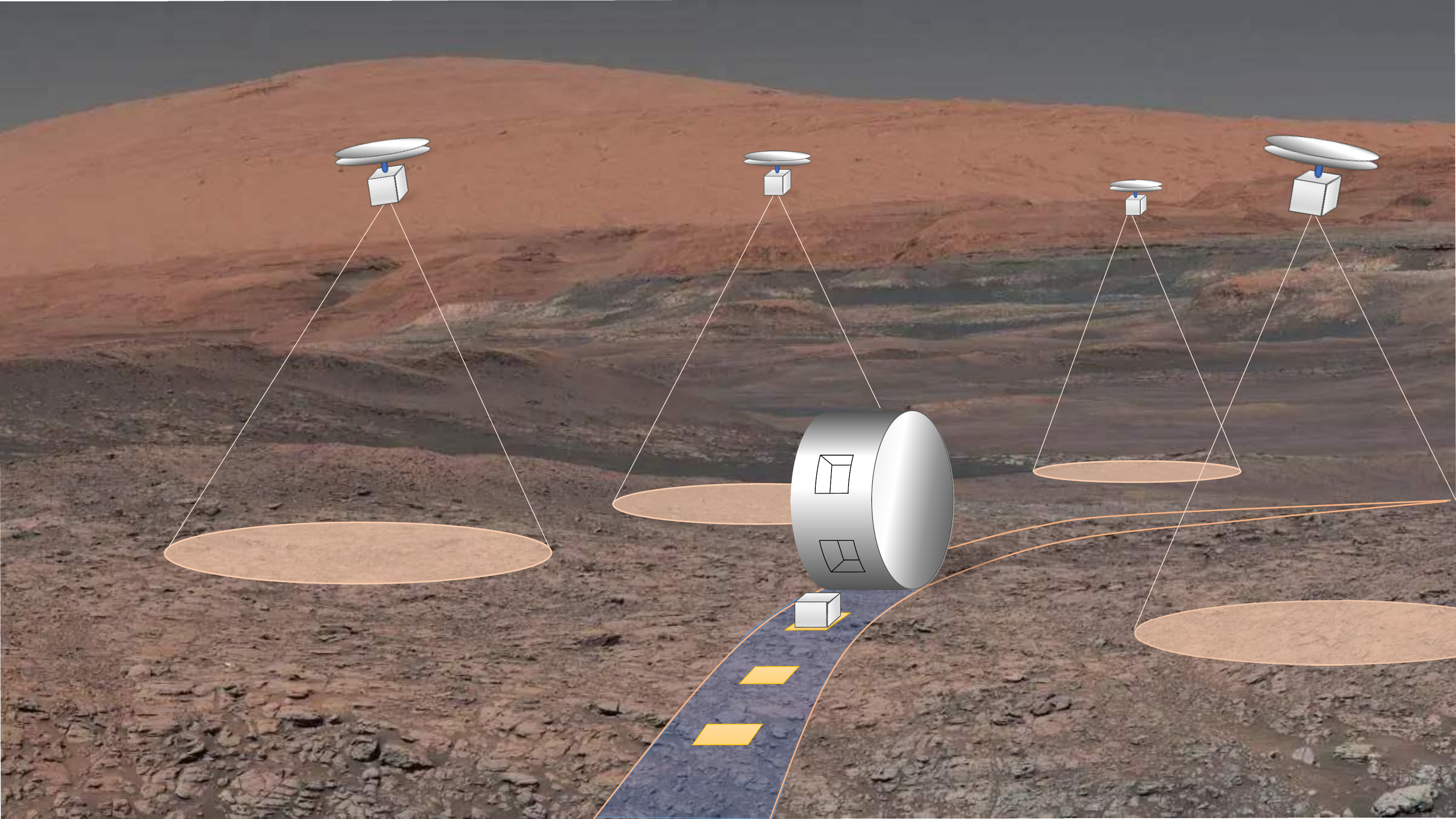}
	\caption{Conceptual layout of our cooperative Mars exploarition system. We evenly arrange several charging slots on the circumference of the rover. Each drone can be recharged by landing on the ground, folding up, and letting the rover roll over it. (The background image was taken by the rover Curiosity.)}
	\label{concept}
\end{figure}

This paper further considers the use of multiple drones for cooperative coverage and designs a novel Mars exploration system (see Fig. \ref{concept}). We strive to maximize the exploration efficiency with the proposed system. However, there are several challenges in its implementation due to the specific characteristics of Mars. For example, the atmospheric pressure of Mars is only about one percent of that on Earth, which greatly increases the energy consumption of drones. Another challenge lies in the large communication latency between Mars and Earth, making it difficult for remote control of distributed drones. Considering the energy limits of drones and requirements for high autonomy \cite{ARCHES}, the simultaneous coverage and charging (SC$^2$) problem becomes our top priority to be dealt with.

The SC$^2$ problem is highly complicated with non-convex constraints aroused by sensor model and safety radius. As such, we solve it with a hybrid algorithmic framework, where cooprative coverage emerges from deep reinforcement learning (DRL) while charging scheduling is achieved by solving a dynamic integer linear programming (ILP). This framework can account for multiple challenges in Mars exploration.

\begin{figure*}
	\includegraphics[width=6.8in]{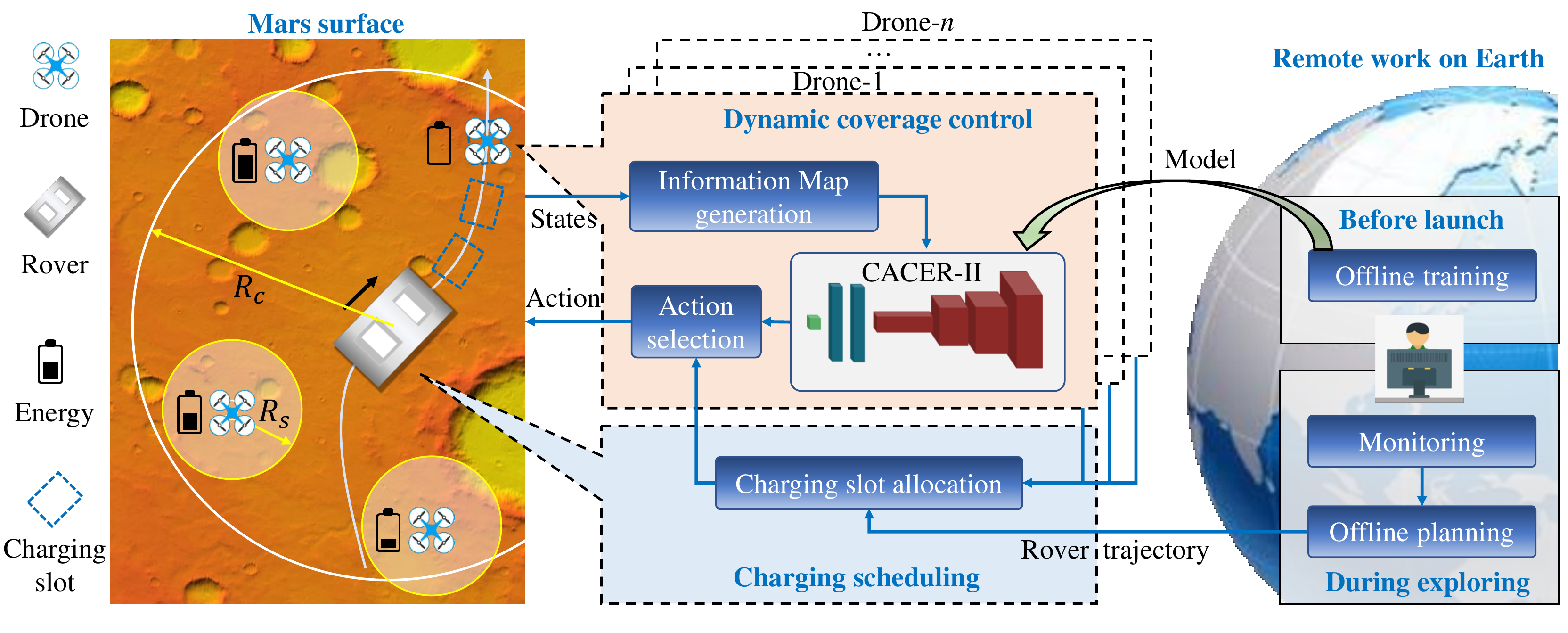}
	\caption{The proposed TIME-SC$^2$ framework. The scenario on the left depicts Mars exploration with the proposed cooperation system. The blocks on the right list key modules for solving the SC$^2$ problem. Our framework establishes a hybrid control architecture with DRL-enabled dynamic coverage control and optimization-based charging scheduling.}
	\label{framework}
\end{figure*}

\subsection{Related work}

As a hot-spot issue, UAV-UGV cooperation has been studied extensively \cite{visibility_constraint,TERRA,ARCHES,MIT}. \cite{visibility_constraint} uses a UAV to explore areas while maintaining visibility to a moving UGV. However, it does not consider the energy limitation. In \cite{TERRA,ARCHES,MIT}, UGV works as a charging station for UAVs, but the charging station is stationary. Our task is more challenging than existing works as multiple energy-limited drones are involved, and the positions of charging stations are time-varying. 

As to field coverage, \cite{Liu_coverage} designs a coverage controller for a group of unicycle-type agents with constant speeds. However, it does not consider the sensor model. \cite{Chen_coverage} assumes that each drone senses a circular area, but the covered area of the fleet over a period of time is not maximized. Besides, anti-flocking algorithms have been used for coverage control in \cite{2016_TII,Ganganath_2018_ISCAS}. However, these algorithms ignore the non-linear properties of the sensor model by assuming that the perception is consistent. Recently, DRL-based coverage control has been put forward in \cite{RL_Coverage,Pham}, but our problem is more intractable since the number of exploring drones and the area of interest change dynamically.

\subsection{Statement of contributions}

This paper presents an overall framework TIME-SC$^2$ for efficient and persistent Mars exploration. Algorithmically, we decompose the highly challenging SC$^2$ problem and solve it stage-wise with two components. First, we design a dynamic coverage algorithm for distributed drones based on deep reinforcement learning. The perception and obstacle information are combined to construct information maps, which maximize the effective coverage of the fleet through time. Then, a charging scheduling algorithm is designed to navigate each drone to a charging slot in need. Experimentally, we present simulations to demonstrate the remarkable performance of TIME-SC$^2$. Our system can explore an area of up to 0.65$\, \rm{km^2}$ within only 1 hour by a 10-drones fleet. Meanwhile, 6 drones will exert the maximum exploration capability with the most streamlined configuration. Moreover, the proposed algorithm is adaptive to diverse tasks that correspond to uneven exploration values of the Mars surface without retraining.

\section{System Model and Problem Formulation}
\label{sec2}
\subsection{System model}
As shown in Fig. \ref{framework}, we consider a cooperation exploration system where $\Lambda_a:=\left \{a_1, ..., a_n\right \}$ drones of a constant speed $v_a \in \mathbb{R}^+$ are employed for simultaneous coverage. Let $p_i(t) \in \mathbb{R}^2$ and $p_r(t) \in \mathbb{R}^2$ be the time-varying positions of $a_i$ and the rover, respectively. The drones need to maintain air-ground connectivity, i.e., $\left\| p_i(t) - p_r(t) \right\| \leq R_c$, where $R_c\in \mathbb{R}^+$ is the communication range. Meanwhile, collision avoidance between drones should be considered, i.e., $\left\| p_i(t) - p_j(t) \right\| \geq R_o, i\neq j$, where $R_o$ is the safety radius. We assume that the rover moves along a pre-defined path $\zeta_r$ with a constant speed $v_r\in \mathbb{R}^+$ and that $v_a>v_r$ since drones are more agile than the rover.

Denote the endurance and consuming time of $a_i$ as $t_a$ and $t_i$, respectively. $a_i$ must return to a charging slot before $t_i$ approaches $t_a$. Specifically, we have a sequence of charging slot candidates $\Lambda_\tau=\left \{\tau_1, ..., \tau_{n_{\tau}}\right \}$ along $\zeta_r$ with $d_\tau$ as the distance between two adjacent slot positions. The number of candidate slots is fixed as $n_\tau=\min\{\frac{R_c}{d_\tau},\frac{t_av_a}{d_\tau}\}$. $n_\tau>n$ can be ensured through delicate design of the rover. The operating mode of $a_i$ is $m_i(t) \in \left \{Explore, Return, Charge\right \}$. Then, starting from $m_i(0)$, the switching function of $m_i(t)$ is given by $m_i(t)=\mathcal{F}(p_i,t_i,x_{ik})$, where $x_{ik}$ is a binary variable that indicates $a_i$ is assigned to $\tau_k$ if $x_{ik}=1$, and 0 otherwise. Note that each drones must be assigned to a charging slot, i.e., $\sum\nolimits_{k=1}^{n_\tau}x_{ik}(t)=1, \, \forall i$. Each charging slot $\tau_k$ can be assigned at most once, i.e., $\sum\nolimits_{i=1}^{n}x_{ik}(t) \leq 1, \, \forall k$. Moreover, the charging slot $\tau_k$ is reachable for $a_i$ only when $\frac{d_{ik}(t)}{v_a}x_{ik}(t) \leq \min \left \{kt_\tau, t_a-t_i \right \}, \; \forall i, k$, where $d_{ik}(t)=\left\| p_i(t) - p_{\tau_k} \right\|$.

The sensor model of $a_i$ is defined by a hill-shaped coverage function as proposed in \cite{sensor_model}:
\begin{equation}
e_i=
\begin{cases}
\frac{M_a}{R_s^4}(c_i^2-R_s^2)^2, \ {\rm if} \; c_i^2 \le R_s \wedge m_i \neq Charge, \\
0, \qquad \qquad \qquad \qquad \ \ \; {\rm else}, 
\end{cases}
\label{sensor}
\end{equation}
where $c_i=\left\| p_i - q \right\|$. $q\in \mathbb{R}^2$ is the position of discretized sensed point. $M_a$ is the peak value when $p_i = q$. $R_s$ is the sensing range. Considering the time-effectiveness of sensor measurements, the effective coverage of $a_i$ at time step $t$ is defined by
\begin{equation}
\tilde e_i(t)= \max \{\tilde e(t-1)-{M_a}/{\eta}, e_i(t) \},
\label{sensor_tail}
\end{equation}
where $\eta$ is a decay factor that characterizes the loss rate of information over time. Then, given $T$ as the mission duration and $A$ as the mission area, the accumulated effective coverage of the drone fleet can be expressed by
\begin{equation}
E=\frac{1}{T}\sum_{i=1}^{n}\sum_{t=t_0}^{T}\sum_{q\in A}{ \tilde e_i(t)}.
\label{effective_coverage}
\end{equation}

\subsection{Problem formulation}

Let $\textbf{P}=\{p_i(t), \forall i,t\}$, $\textbf{X}=\{x_{ik}(t), \forall i,k,t\}$. Our objective is to maximize the accumulated effective coverage $E$ by jointly optimizing drone trajectories (i.e., $ \textbf{P} $) and charging scheduling (i.e., $ \textbf{X} $). The optimization problem is formulated as
\begin{subequations}
	\label{optimization}
	\begin{align}
	& \max_{\textbf{P},\textbf{X}}\quad E \\
	& {\;\rm s.t.} \quad \; p_i(t)=p_i(t-1)+v, \qquad \forall i,t, \label{dynamics}\\
	& \qquad \;\;\, \left\| p_i(t) - p_r(t) \right\| \leq R_c, \qquad \forall i,t, \label{comm}\\
	& \qquad \;\;\, \left\| p_i(t) - p_j(t) \right\| \geq R_o,  \qquad \forall i,j,t,i\neq j, \label{collision}\\
	& \qquad \;\;\, \sum\nolimits_{k=1}^{n_\tau}x_{ik}(t)=1, \; \qquad \quad \, \forall i,t,  \label{eq:2B}\\
	& \qquad \;\;\, \sum\nolimits_{i=1}^{n}x_{ik}(t) \leq 1, \; \qquad \quad \;\forall k,t,  \label{eq:2C}\\
	& \qquad \;\;\, \frac{d_{ik}(t)}{v_a}x_{ik}(t) \leq \min \left \{kt_\tau, t_a-t_i \right \}, \;  \forall i,k,t,  \label{eq:2D}\\
	& \qquad \;\;\, x_{ik}(t) \in \left \{0,1\right \}, \qquad \qquad \; \; \;\forall i, k,  \label{eq:2E}
	\end{align}
	\nonumber
\end{subequations}
where $\left\| v \right\|=v_a$. The formulated problem is a mixed integer non-convex optimization problem that is challenging to solve.

\section{Approach}

\label{sec3}

In this section, we propose a hybrid algorithmic framework for solving the coupled SC$^2$ problem (\ref{optimization}). We note that the two sub-problems of SC$^2$, namely coverage control and charging scheduling, can be handled in different ways due to their unique formulations. First, the charging scheduling can be reorganized into a standard ILP, which is therefore solvable. Second, we resort to DRL for acquiring optimal coverage control policies through agent-environment interaction. By such we can avoid the difficulties aroused by non-convex constraints in problem (\ref{optimization}). 

As to coverage control, we construct an information map for joint state representation and design a novel learning algorithm that maps the information map to optimal actions. Also, we propose a charging scheduling algorithm that involves charging slot allocation and mode switching as key supplements.

\subsection{Information maps}

\begin{figure}
	\centering
	\includegraphics[width=3.3in]{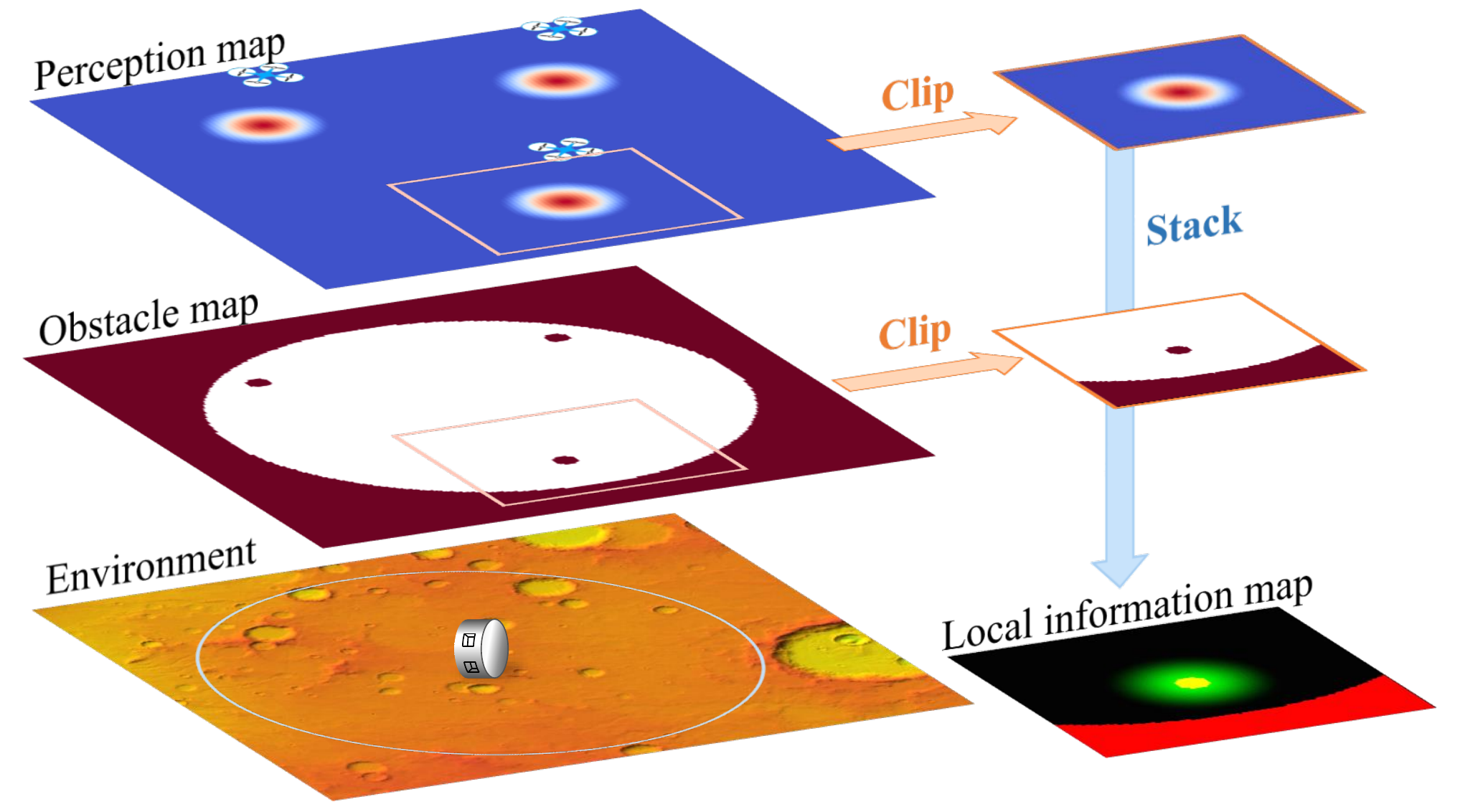}
	\caption{An example for an information map that is stacked by a perception map and a obstacle map. The state representation is a local information map centered on each drone with size $2R_s \times 2R_s$.}
	\label{informap}
\end{figure}

We introduce a novel version of the information map $I$, which is stacked by a perception map and an obstacle map with the same size, as depicted in Fig. \ref{informap}. The perception map records sensing history of drones, while the obstacle map defines the feasible area for drones.

Let $M$ be the perception map centered on $p_r$ with side length $2(R_s+R_c)$, which is then discretized into a set of square cells with size $c$. Each cell in $M$ is denoted by $M(q)$ where $q$ is the center coordinate of the cell and let $Q$ be a set of all such $q$ values within $M$. Here, $M(q)$ reflects how effective the drone senses a point $q$, which decays as time evolves. At the beginning, the perception map is initialized with $M(q) = 0$ for all $q \in Q$. In consistent with (\ref{sensor_tail}), $M$ is updated by
\begin{equation}
M(q) \leftarrow \max \left \{ {M(q)-{M_a}/{\eta},\; {E_a}\left( {{{
				p}_i},\;{q}} \right)} \right \}, \ \forall i.
\label{update}
\end{equation}

Note that the perception map will shift with the movement of the rover. Let $p'_r$ be the position of the rover at the next time step, the shift rule of the perception map is defined by $M(q) \leftarrow M(q+\Delta q)$ for all cells that belong to the intersection of the new perception map and the old map, where $\Delta q = p'_r-p_r$.

Now we consider the construction of the obstacle map to represent the positions of all drones and the communication range of the rover. Areas beyond the communication range are equivalently regarded as obstacles. As such, the obstacle map $O$, also depicted in Fig. \ref{informap}, is created by:
\begin{equation}
O(q) = \left\{ {\begin{array}{*{20}{c}}
	1,&{{\rm{if}} \ \left\| {{{p}_i} - {q}} \right\| < {R_o} \ {\rm{
				or}}\;\left\| {{q}} \right\| > {R_c}},\\
	0,&{{\rm{otherwise}}}.
	\end{array}} \right.
\label{obstacle}
\end{equation}

\subsection{CACER-II algorithm}

We deal with the multi-drone coverage control  problem in the context of DRL. The individual state representation $s$ is a local receipt of the information map with size $2R_s \times 2R_s$ centered on each drone. Such segmentation abandons the negligible impact of environment information in the distance and reduces the communication traffic. 

We steer the movement of each drone by adjusting its heading $\psi_i \in [ - \pi ,\;\pi) $. This is more general than \cite{RL_Coverage}, which assumes that the action space is discretized. 

The reward function $r$ is specified as $r = \omega_c{r_c} + \omega_e{r_e} + {r_p}$, where $r_c$, $r_e$, and $r_p$ represent coverage reward, exploration reward, and collision penalty, separately. $\omega_c$ and $\omega_e$ are tuning parameters. Specifically, the coverage reward ${r_c}$ is defined as
\begin{equation}
{r_c} = \frac{{\sum M  - n\sum {{M_0}} }}{{n\sum {{M_0}} }},
\label{Rc}
\end{equation}
where ${M_0}$ is the baseline perception map containing only one stationary drone. To encourage the drone to explore new areas, the exploration reward ${r_e}$ is designed as
\begin{equation}
{r_e} = \sum {M'}  - \sum M ,
\label{Re}
\end{equation}
where $M'$ is the new perception map updated with the sensor information of the current decision-making drone. Besides, we design a penalty item ${r_p}$ to enforce drones to stay within the feasible area as follows:
\begin{equation}
{r_p} = \left\{ {\begin{array}{*{20}{c}}
	{ - \left\| {{p_i} - {p_r}} \right\|},&{{\rm{if}}\left\| {{p_i} - {p_r}} \right\|> {R_c}},\\
	{ - 200},&{{\rm{if}}\;\left\| {{p_i} - {p_j}} \right\| < {R_o}},\\
	0,& \rm{otherwise},\\
	\end{array}} \right.
\end{equation}
for all $j \ne i$.

To solve the formulated RL problem, we propose a novel DRL algorithm CACER-II, which is an extension of our previous work, \textit{Continuous Actor-Critic with Experience Replay} (CACER) \cite{Yan}. In this paper, we have redesigned and improved CACER throughly to meet the challenges in multi-drone scenarios (see Algorithm \ref{alg1}). As opposed to CACER that controls a single agent, CACER-II is able to optimize multi-drone coverage control in a nonstationary environment. This is achieved by learning from multiple drones' experiences and sharing the same policy among the homogeneous fleet. We note that CACER-II follows a centralized-learning and decentralized-execution paradigm \cite{AAMAS}. During the execution stage, each drone selects its own action with locally perceived states in a fully decentralized manner. Such paradigm gives CACER-II the scalability to different number of drones.

\begin{algorithm}[t!]
	\caption{Simultaneous Coverage with CACER-II}
	\renewcommand{\algorithmicrequire}{\textbf{Initialize}}
	\renewcommand{\algorithmicensure}{\textbf{Output:}}
	\label{alg1}
	\begin{algorithmic}[1]
		\REQUIRE replay memory $D$, actor network $Act\left( {s|\,{\theta ^A}} \right)$, critic network $V\left( {s|\,{\theta ^V}} \right)$, learning rate $\alpha$ and $\beta$, Gaussian noise $\mathcal{N}\sim N(0,\sigma^2)$, training batch size $N_b$
		\FOR {episode = 1, 2, ...}
		\STATE Generate $n$, $p_r$, $p_i$ and $\zeta_r$ randomly
		\STATE Construct information map $I\leftarrow {\rm Stack} {(M,O)}$
		\STATE Represent the observed states $s_i\leftarrow {\rm Clip}{(I)}$
		\FOR {$t$ = 1 to $t_a$}
		\FOR {each $a_i \in \Lambda_a$}
		\STATE ${\psi_i} \leftarrow Act\left( {s_i|{\theta ^A}} \right) + \mathcal{N}$
		\STATE $p'_{ix} \leftarrow p_{ix} + v_a \cos(\pi \psi_i)$; $p'_{iy} \leftarrow p_{iy} + v_a \sin(\pi \psi_i)$
		\ENDFOR
		\STATE Update information map $I$
		\FOR {each $a_i \in \Lambda_a$}
		\STATE Clip $s'_i$ from $I$
		\STATE Calculate immediate reward ${r_i}$
		\STATE Store tuple $({s_i},{\psi_i},{r_i},s'_i)$ in $D$
		\STATE $s_i \leftarrow s'_i$; $p_i \leftarrow p'_i$
		\ENDFOR
		\STATE Sample $N_{b}$ tuples $({s_k},{\psi_k},{r_k},s'_k)$ from $D$
		\STATE Calculate temporal-difference error: \\
		${\delta _k} = {r_k} + \gamma  \cdot V\left( {{{s'}_k}|{\theta ^V}} \right) - V\left( {{s_k}|{\theta ^V}} \right)$
		\STATE Optimize ${\theta ^A}$ if ${\delta _k} > 0$: \\
		$\theta^{A} \leftarrow \theta^{A}+ \alpha (\psi_{k}-A c t (s_{k} | \theta^{A})) \frac{\partial A c t(s_{k} | \theta^{A})}{\partial \theta^{A}}$
		\STATE Optimize ${\theta ^V}$ with $\theta^{V} \leftarrow \theta^{V}+\beta \delta_{k} \frac{\partial V(s_{k} | \theta^{V})}{\partial \theta^{V}}$
		\ENDFOR
		\ENDFOR
	\end{algorithmic}
\end{algorithm}

The network structure of CACER-II is briefly introduced as follows. Two deep neural networks with the same structure are used to represent the actor and the critic, respectively. We use local information map as the input, which is resized into 84 $\times{}$ 84 $\times{} $ 2. The input is successively passed by four convolutional layers with the ReLU activation function: The first convolutional layer convolves 32 filters of 8 $\times{}$ 8 with stride 4, followed by the second one with 64 filters of 4 $\times{}$ 4 with stride 2. The third one has 64 filters of 3 $\times{}$ 3 with stride 1, followed by the last one with 512 filters of 7 $\times{}$ 7 with stride 1. After that, the output of the convolutional neural networks is flattened and then fed to two fully-connected layers with 256 hidden units, ReLU activation function. Note that the actor uses a hyperbolic tangent activation function, while the critic uses a linear activation function.
\begin{figure}
	\centering
	\includegraphics[width=3.3in]{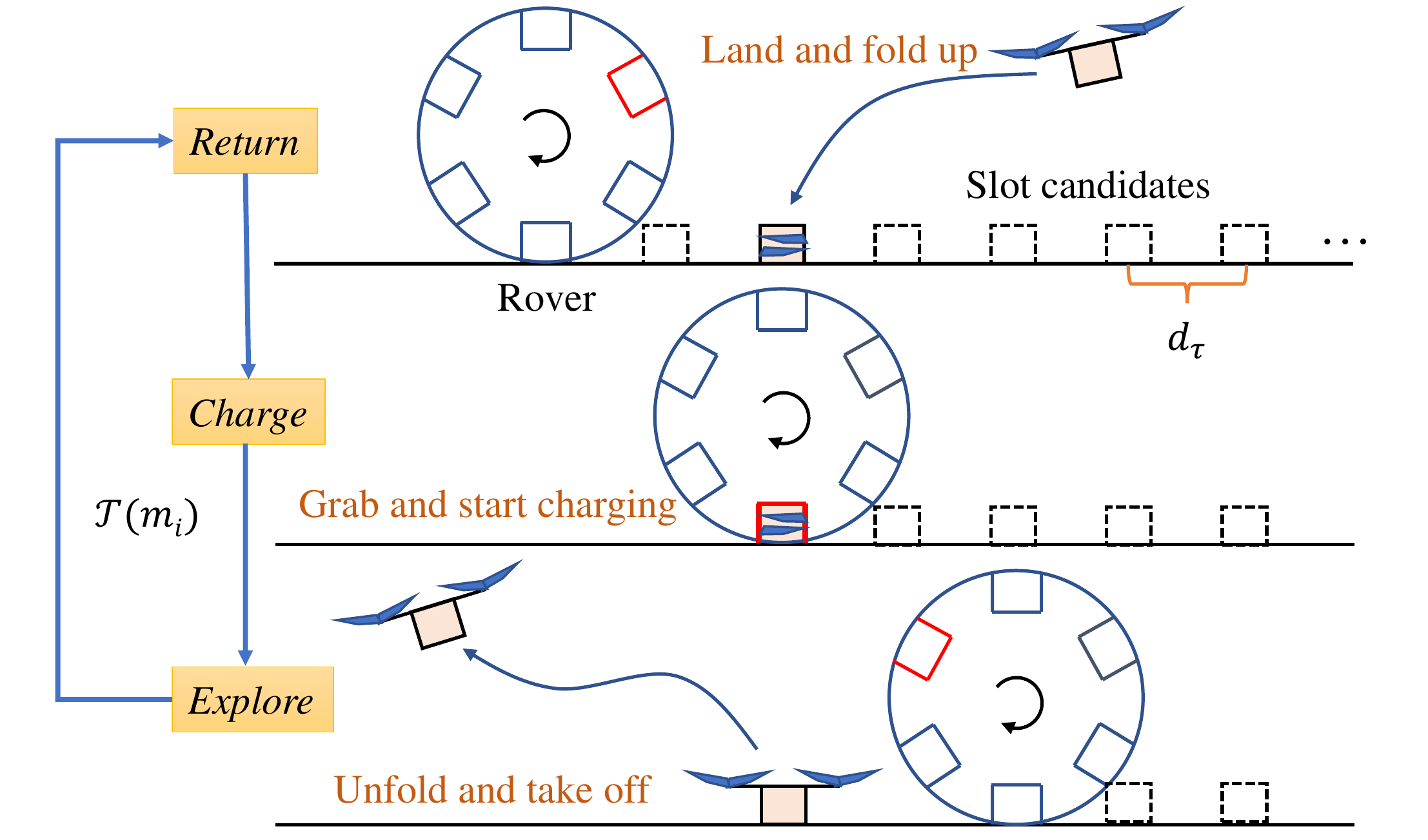}
	\caption{Illustration of the charging process and switching between different operating modes.}
	\label{stages}
\end{figure}

\begin{figure*}
	\centering
	\includegraphics[width=6.8in]{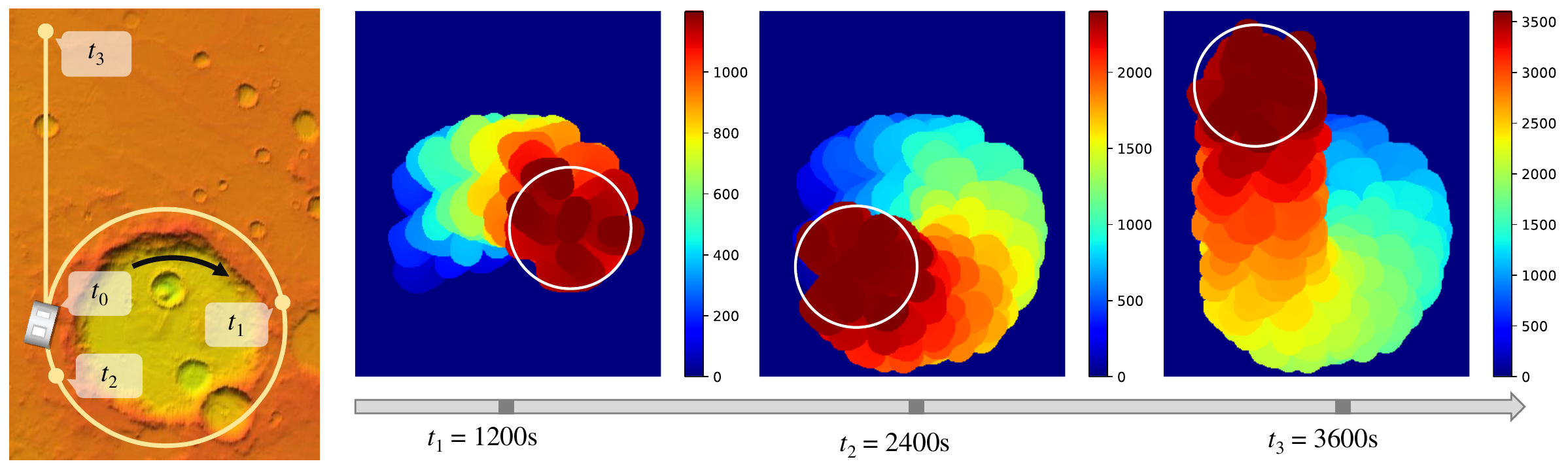}
	\caption{The designed rover path and snapshots of multi-drone coverage areas ($n=10$). The mission begins at $t_0$, where the drones depart from the rover in turn, and ends at $t_6=3600 \, \rm{s}$. The color intensity in each coverage map is positively correlated with time, and we use a white circle to indicate the feasible area for drones.}
	\label{results}
\end{figure*}

\subsection{Action selection}

The switching topology of $m_i(t)$ is a unidirectional ring, as illustrated in Fig. \ref{stages}. Given current states of $a_i$ associated with the assigned charging slot $\tau_k$, the switching function $\mathcal{F}$ is alternatively represented in a transition form as $m_i(t+1)=\mathcal{T}(m_i(t))$, given by
\begin{itemize}
	\item $\mathcal{T}(m_i(t)):Explore\to Return$ if $\left \| p_i(t)-p_{\tau_k} \right \|/v_a>(t_a-t_i-2)$;
	
	\item $\mathcal{T}(m_i(t)):Return\to Charge$ if $\left \| p_i(t)-p_{\tau_k} \right \|<\epsilon$, where $\epsilon$ is a small positive threshold;
	
	\item $\mathcal{T}(m_i(t)):Charge\to Explore$ if $ \left \| p_i(t)-p_r \right \|<\epsilon$.
\end{itemize}

The coverage control during exploration is enabled by CACER-II. Assume that the charging slot $\tau_k$ has been assigned, the drone can autonomously navigate to $\tau_k$ based on current mode $m_i$ determined by $\mathcal{F}$. Above all, the control policy associated with mode switching for cooperative Mars exploration is summarized in Algorithm \ref{alg2}. The next section will complete the charging slot allocation algorithm.

\begin{algorithm}[h!]
	\caption{Action Selection}
	\renewcommand{\algorithmicrequire}{\textbf{Input:}}
	\renewcommand{\algorithmicensure}{\textbf{Output:}}
	\newcommand{\SWITCH}[1]{\STATE \textbf{switch} #1 \textbf{do}}
	\newcommand{\ENDSWITCH}{\STATE \textbf{end switch}}
	\newcommand{\CASE}[1]{\STATE \hspace{11.5pt} \textbf{case} #1 \begin{ALC@g}}
		\newcommand{\ENDCASE}{\end{ALC@g}}
	\newcommand{\CASELINE}[1]{\STATE \textbf{case}}
	\newcommand{\DEFAULT}{\STATE \textbf{default:} \begin{ALC@g}}
		\newcommand{\ENDDEFAULT}{\end{ALC@g}}
	\newcommand{\DEFAULTLINE}[1]{\STATE \textbf{default:}}
	\label{alg2}
	\begin{algorithmic}[1]
		\REQUIRE $-$ Combined states $\{p_i(t), t_i(t), m_i(t)\}$ of each drone; \\
		\hspace{11.5pt} $-$ Global position of the rover $p_r(t)$; \\
		\hspace{11.5pt} $-$ Trajectory of the rover $\zeta_r$ \\
		\ENSURE The control inputs for each drone $\psi_i(t)$
		\STATE Update $m_i(t)$ according to $\mathcal{F}$
		\STATE Solve (\ref{Integer_plan}) for a optimized chargiing scheduling $\tau_k$\\
		\STATE Calculate $p_{\tau_k}$ based on $\zeta_r$ and $\tau_k$
		\SWITCH {$m_i(t)$}
		\CASE {$Explore$}
		\STATE \hspace{11.5pt} {Obtain $\psi_i(t)$ with the learned CACER-II policy}
		\ENDCASE
		\CASE {$Return$}
		\STATE \hspace{11.5pt} {$\psi_i(t)= \arctan \frac{p_{\tau_k}-p_i(t)}{\left \| p_{\tau_k}-p_i(t) \right \|} $}
		\ENDCASE
		\CASE {$Charge$}
		\STATE \hspace{11.5pt} {Set the drone speed to zero}
		\ENDCASE
		\ENDSWITCH
	\end{algorithmic}
\end{algorithm}

\subsection{Charging slot allocation}

Now we consider developing a slot allocation algorithm to ensure persistent execution of Mars exploration. Our goal is to assign each drone $a_i \in \Lambda_a$ to a charging slot $\tau_k \in \Lambda_\tau$ without conflicts. First, if $m_i(t)=Explore$ for all $a_i\in \Lambda_a$, the slot allocation problem at time $t$ can be formulated as
\begin{subequations}
	\label{Integer_plan}
	\begin{align}
	& \min_{x_{ik}} \,\, z= \sum_{k=1}^{n_\tau} \sum_{i=1}^{n} d_{ik}(t) x_{ik}(t) \\
	& {\;\rm s.t.} \quad \text{(\ref{eq:2B}),\; (\ref{eq:2C}),\; (\ref{eq:2D}).} \label{eq:11B}
	\end{align}
\end{subequations}

Since problem (\ref{Integer_plan}) is a standard ILP, it can be solved efficiently by existing optimization tools such as CVX \cite{CVX}. Furthermore, problem (\ref{Integer_plan}) must be adjusted dynamically to adapt to mode switching of multiple drones. As such, we add the following additional constraints to complete problem (\ref{Integer_plan}): for each $a_i \in \Lambda_a$, if $m_i(t)\neq Explore$, then $a_i$ is excluded form the planning. for each $\tau_k \in \Lambda_\tau$, if $\tau_k$ has been occupied by a non-exploring drone $a_i$, then $\tau_k$ is excluded from the planning. The following theorem ensures that problem (\ref{Integer_plan}) has a feasible solution at any time.

\noindent \textbf {Theorem 1:} By applying the proposed charging scheduling algorithm along with the action selection rule, there is always a feasible solution for problem (\ref{Integer_plan}).

\noindent \textbf {Proof:} The idea of our proof is similar to mathematical induction. With appropriate system settings, we can ensure that there is a feasible solution at the begining. Then, we will show that if there exsits a feasible solution at time $t_0 > 0$, there is at least a feasible solution for $t_1=t_0+1$.

For any $a_i$, the possibilities for mode trasition $\mathcal{T}(m_i)$ from $t_0$ to $t_1$ can be partitioned into the following three cases.

\noindent \textit {Case 1:} $\mathcal{T}(m_i):Explore\to Explore$. Consider the worst case, where $a_i$ moves in the opposite direction of the assigned charging slot $\tau_k$. Since $\left \| p_i(t_0)-p_{\tau_k} \right \|/v_a<(t_a-t_i-2)$, we have $\left \| p_i(t_1)-p_{\tau_k} \right \|/v_a<(t_a-t_i)$, which means $\tau_k$ is still reachable for $a_i$.

\noindent \textit {Case 2:} $\mathcal{T}(m_i) \in \{Explore\to Return, Return\to Return, Return\to Charge, Charge\to Charge\}$. It is easy to verify that the assigned slot at time $t_0$ is still reachable for $a_i$ at time $t_1$, which is omitted here due to page limits.

\noindent \textit {Case 3:} $\mathcal{T}(m_i):Charge\to Explore$. We have $t_i(t_1)=0$, which means any free charging slot is reachable for $a_i$.

Above all, we can conclude that there is at least a feasible solution for problem (\ref{Integer_plan}) for any $t \geq 0$. \qed

\begin{figure}[t!]
	\includegraphics[width=3.2in]{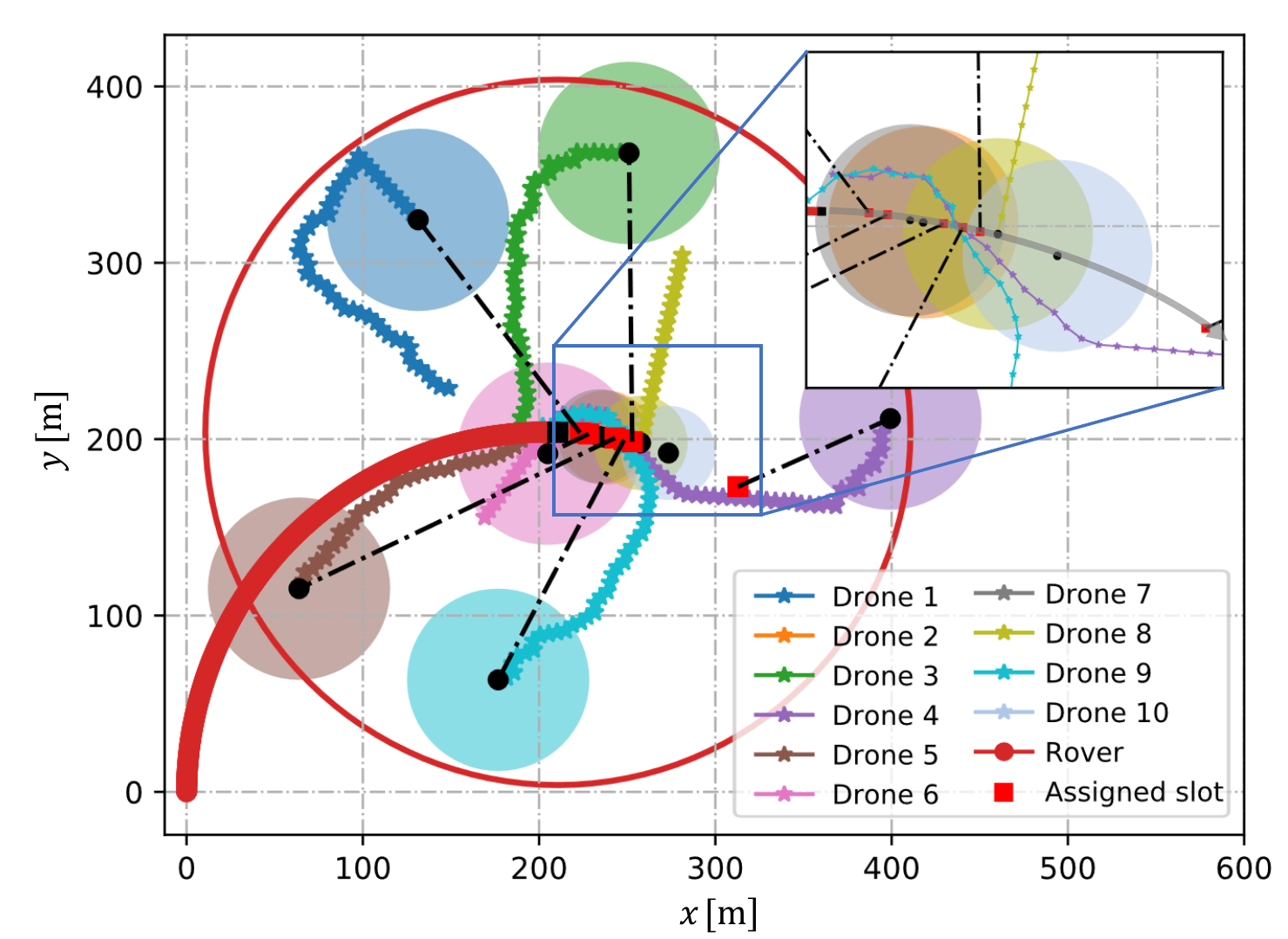}
	\caption{A snapshot of drone trajectories at $t=650 \,\rm{s}$. The dashed lines are used to connect drones with currently assigned charging slots. We use large circles to represent drones under exploration, and small circles to represent drones parked in the charging slots.}
	\label{trajectory}
\end{figure}

\section{Results and Discussions}

This section demonstrates the efficiency of TIME-SC$^2$ under typical Mars exploration missions, investigates the evolution of exploration efficiency with respect to fleet size, and verifies its adaptability to the Mars surface. The experimental settings are given in Table I.

\begin{table}[htbp]
	\centering
	\captionsetup{justification=centering}
	\caption*{TABLE I: Experimental parameters.}
	\begin{tabular}{cc||cc}
		\toprule
		\multicolumn{2}{c}{Drone / Rover} & \multicolumn{2}{c}{CACER-II}\\
		\midrule
		Drone Speed $v_a$ & 5 m/s & Learning rate $\alpha$ & $10^{-4}$ \\
		Sensor range $R_s$ & 50 m & Learning rate $\beta$ & $10^{-3}$\\
		Endurance $t_a$ & 100 s & Reward weight $\omega_c$ & 20\\
		Safety radius $R_o$ & 5 m & Reward weight $\omega_e$ & 1 \\
		Rover Speed $v_r$ & 0.5 m/s & Gaussian noise $\sigma$ & $0.5\to 0.05$\\
		Comm. range $R_c$ & 200 m & Training batch size $N_b$ & 64 \\
		Slot interval $d_\tau$ & 5 m & Discounted factor $\gamma$ & 0.95\\
		\bottomrule
	\end{tabular}
\end{table}

\subsection{Perform a typical Mars exploration mission}

The designed rover path consists of two stages: i) first circle around a crater with a radius of $200 \, \rm{m}$, and ii) then explore $500 \, \rm{m}$ in a straight line. Fig. \ref{results} provides a temporal illustration of our mission. As shown, the coverage area expands over time. By the end of the mission, an area up to $0.65 \, \rm{km^2}$ has been explored within only $3600 \, \rm{s}$, which is $98.44\%$ of the feasible area constrained by the communication range. Besides, it shows satisfactory continuity in the evolution of the coverage area despite the energy limits of drones.

A snapshot of the drone trajectories at $t=650 \, \rm{s}$ is shown in Fig. \ref{trajectory}. We notice that $6$ drones in $Explore$ mode are evenly distributed in space. They tend to explore unfamiliar areas, move apart from neighbors, and preserve connectivity to the rover. This result shows that we can achieve the predetermined control objectives through the learned CACER-II policy. Moreover, each drone is assigned to a charging slot, which indicates that the proposed charging scheduling algorithm ensures conflict-free real-time slot allocation.

\subsection{Correlation between efficiency and number of drones}

In order to quantitatively evaluate the exploration efficiency with different number of drones, we define the cumulative coverage ratio $\Gamma_{\rm{cum}}$ as 
\begin{equation}
\Gamma_{\rm{cum}} = \frac{\sum\limits_{q \in A} \max\limits_{t}M(q,t)}{\sum\limits_{q \in A} M_1(q)},
\label{CumulativeCoverage}
\end{equation}
where $M_1$ is the perception map with all 1 values, $A$ is the feasible area throughout the mission. For comparison, we also define the average instantaneous coverage ratio $\Gamma_{\mathrm{avg}}$ as
\begin{equation}
\Gamma_{\rm{avg}} = \frac{1}{T}\sum\limits_{t}\frac{\sum\limits_{q\in A_0} M(q,t)}{\sum\limits_{q\in A_0} M_1(q)},
\label{InstantCoverage}
\end{equation}
where $A_0$ is the circular feasible area at a certain moment. $T$ is the mission duration.

\begin{figure}
	\centering
	\includegraphics[width=3.3in]{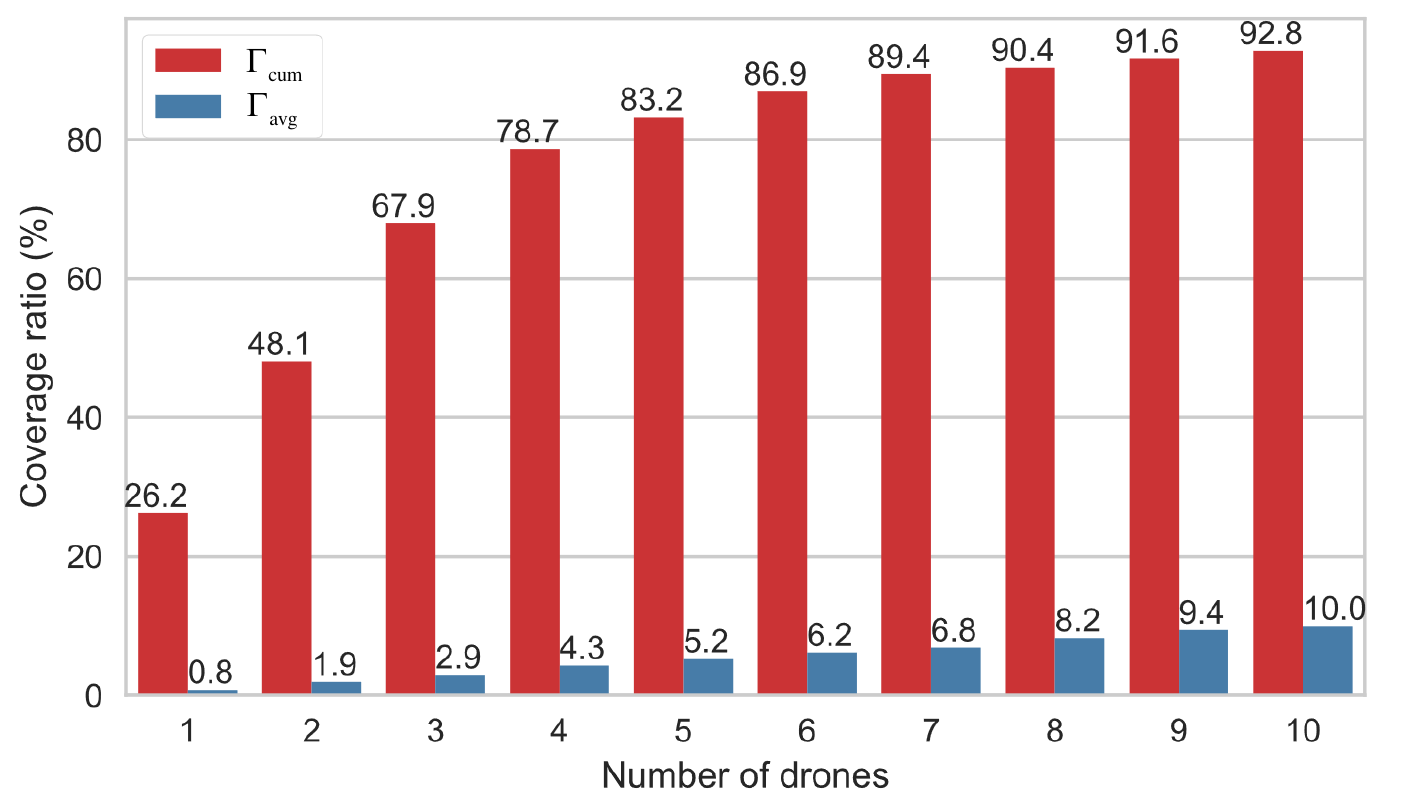}
	\caption{Comparison of the cumulative coverage ratio $\Gamma_{\rm{cum}}$ and the average  instantaneous coverage ratio $\Gamma_{\rm{avg}}$ cunder different numbers of drones.}
	\label{boxplot}
\end{figure}

A series of experiments have been conducted with different number of drones. In all the experiments, the rover moves $1000 \, \rm{s}$ in a straight line. As depicted in Fig. \ref{boxplot}, both $\Gamma_{\rm{cum}}$ and $\Gamma_{\rm{avg}}$ increase with the number of drones. $\Gamma_{\rm{cum}}$ is much higher than $\Gamma_{\rm{avg}}$, which indicates that through the proposed TIME-SC$^2$ the exploration efficiency has been significantly improved by cooperation of multiple drones. Besides,
the growth of $\Gamma_{\rm{cum}}$ has slowed down after the number of drones exceeds 6, where we have $\Gamma_{\rm{cum}}>85\%$. Such performance is satisfactory since the cumulative coverage ratio has taken the sensor model into consideration. Therefore, we can conclude that under the current parameter settings, 6 drones will exert the maximum exploration capability of the system with the most streamlined configuration. This conclusion provides meaningful guidance for the future applications.

\begin{figure}
	\centering
	\includegraphics[width=3.3in]{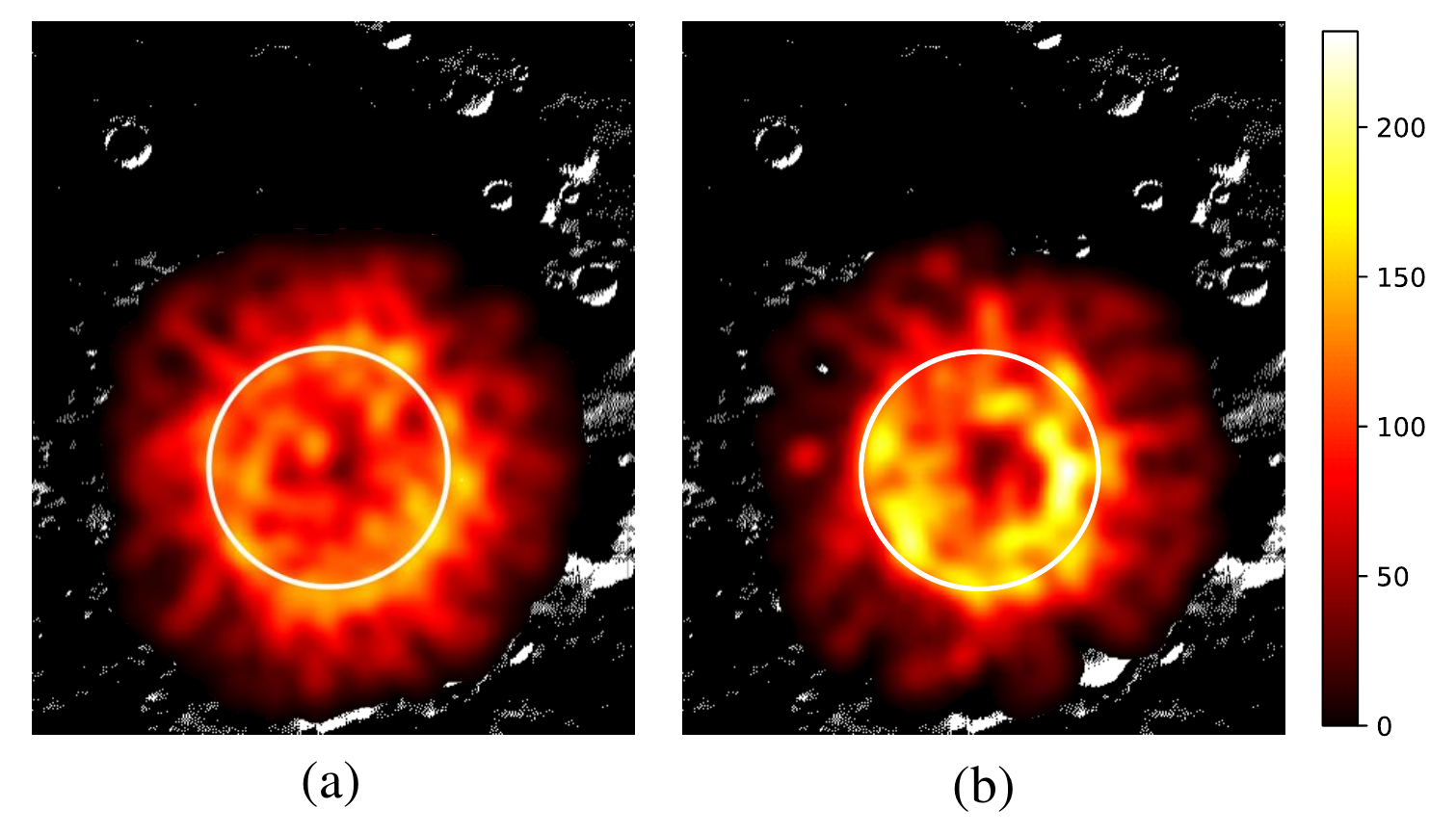}
	\caption{Comparison between environment-independent exploration (a) and environment-adaptive exploration (b). The areas within the while circle is a crater, which deserves detailed exploration.}
	\label{adaptive}
\end{figure}

\subsection{Adaptability to the Mars surface}
The exploration value of the Mars surface is uneven in different areas. Therefore, the drones should be steered to explore high-value areas more, such as a crater. This is accomplished by modifying information maps without retraining of the CACER-II network. For example, the map of the mission area can be preprocessed through a saliency detection algorithm \cite{SR}. Then, the texture layer is fused into the information map as inputs. A comparison between environment-independent exploration and environment-adaptive exploration is depicted in Fig. \ref{adaptive}. By considering the texture of Mars surface, the drone fleet tend to explore the crater more. This result demonstrates the adaptivity of TIME-SC$^2$.

\section{Concluding Remarks}

In this paper, we have designed a comprehensive framework for cooperative Mars exploration. A series of simulations have been conducted to demonstrate its remarkable time-efficiency, scalability to the number of drones, and adaptivity to the Mars surface. Note that the structure of the rover is not specified. The proposed TIME-SC$^2$ framework is general for systems with similar configurations.

A meaningful future direction lies in real-time path planning for the rover with the assistance of drones, thereby avoiding potential dangers such as sand traps \cite{Spirit,Hybrid}. Another promising direction is to extend exploration areas by establishing communication links between drones \cite{JSAC}, which requires a fully distributed control architecture.

\addtolength{\textheight}{-3cm}   






\begin{thebibliography}{99}

\bibitem{Roadmap} International Space Exploration Coordination Group, ``The global exploration roadmap,'' Washington: NASA, 2018.

\bibitem{Spirit} K. C. Di and Z. J. Ge, ``A brief review of Spirit's six years of Mars roving and scientific discoveries,'' \textit{Journal of Remote Sensing}, vol. 15, no. 4, pp. 651--658, 2011.

\bibitem{mars_2020} Y. Zheng, ``Mars Exploration in 2020,'' \textit{Innovation}, vol. 1, no. 2, p. 100036, 2020.

\bibitem{TERRA} R. Fernando, P. Munoz, and M. D. R-Moreno, ``TERRA: A Path Planning Algorithm for Cooperative UGV-UAV Exploration,'' \textit{Engineering Applications of Artificial Intelligence}, vol. 78, pp. 260--272, 2019.

\bibitem{ARCHES} M. J. Schuster, et al., ``The ARCHES Space-Analogue Demonstration Mission: Towards Heterogeneous Teams of Autonomous Robots for Collaborative Scientific Sampling in Planetary Exploration,'' \textit{IEEE Robotics and Automation Letters}, vol. 5, no. 4, pp. 5315--5322, 2020.

\bibitem{MIT} M. Valenti, B. Bethke, J. P. How, D. P. de Farias and J. Vian, ``Embedding Health Management into Mission Tasking for UAV Teams,'' in \textit{2007 American Control Conference}, New York, 2007, pp. 5777--5783.

\bibitem{visibility_constraint} K. Lukas, S. Khodaverdian, and V. Willert, ``Motion control for UAV-UGV cooperation with visibility constraint,'' in \textit{2015 IEEE Conference on Control Applications (CCA)}, pp. 1378--1385, 2015.

\bibitem{Liu_coverage} Q. Liu, M. Ye, Z. Sun, J. Qin, and C. Yu, ``Coverage Control of Unicycle Agents under Constant Speed Constraints,'' in \textit{IFAC World Congress}, vol. 50, no. 1, pp. 2471--2476, 2017.

\bibitem{Chen_coverage} R. Chen, N. Xu, and J. Li, ``A Self-Organized Reciprocal Decision Approach for Sensing Coverage with Multi-UAV Swarms,'' in \textit {Sensors}, vol. 18, no. 6, p. 1864, 2018.

\bibitem{2016_TII} N. Ganganath, C. Cheng and C. Tse, ``Distributed Antiflocking Algorithms for Dynamic Coverage of Mobile Sensor Networks,'' in \textit{IEEE Transactions on Industrial Informatics}, vol. 12, no. 5, pp. 1795--1805, 2016.

\bibitem{Ganganath_2018_ISCAS} N. Ganganath, W. Yuan, C. Cheng, T. Fernando and H. H. C. Iu,``Territorial marking for improved area coverage in anti-flocking-controlled mobile sensor networks'' in \textit{IEEE International Symposium on Circuits and Systems (ISCAS)}, 2018, pp. 1--4.

\bibitem{RL_Coverage} Xiao, Jian, et al., ``A Distributed Multi-Agent Dynamic Area Coverage Algorithm Based on Reinforcement Learning,'' \textit{IEEE Access}, vol. 8, pp. 33511--33521, 2020.

\bibitem{Pham} H. X. Pham, H. M. La, D. Feil-Seifer, A. Nefian, ``Cooperative and distributed reinforcement learning of drones for field coverage,'' \textit{arXiv preprint arXiv:1803.07250,} 2018. 

\bibitem{sensor_model} I. I. Hussein and D. M. Stipanovi, ``Effective Coverage Control for Mobile Sensor Networks With Guaranteed Collision Avoidance,'' \textit{IEEE Transactions on Control Systems Technology}, vol. 15, no. 4, pp. 642--657, 2007.

\bibitem{Yan} C. Yan, X. Xiang, and C. Wang, ``Fixed-Wing UAVs flocking in continuous spaces: A Deep reinforcement learning approach,'' \textit{Robotics and Autonomous Systems}, vol. 131, p. 103594, 2020.

\bibitem{AAMAS} K. G. Jayesh, E. Maxim, and K. Mykel, ``Cooperative multi-agent control using deep reinforcement learning'', in \textit{International Conference on Autonomous Agents and Multiagent Systems (AAMAS)}, pp. 66--83, 2017

\bibitem{CVX} M. Grant and S. Boyd, \textit{CVX: MATLAB Software for Disciplined Convex Programming,} 2016. Available: http://cvxr.com/cvx

\bibitem{SR} X. Hou and L. Zhang, ``Saliency Detection: A Spectral Residual Approach.'' in \textit{IEEE Conference on Computer Vision \& Pattern Recognition (CVPR)}, pp. 18--23, 2007.
 
\bibitem{Hybrid} J. Li, G. Deng, C. Luo, Q. Lin, Q. Yan and Z. Ming, ``A Hybrid Path Planning Method in Unmanned Air/Ground Vehicle (UAV/UGV) Cooperative Systems,'' \textit{IEEE Transactions on Vehicular Technology}, vol. 65, no. 12, pp. 9585--9596, 2016.

\bibitem{JSAC} C. Liu, Z. Chen, J. Tang, J. Xu, C. Piao, ``Energy-Efficient UAV Control for Effective and Fair Communication Coverage: A Deep Reinforcement Learning Approach,'' \textit{IEEE Journal on Selected Areas in Communications,} vol. 36, no. 9, pp. 2059--2070, 2018.

\end{thebibliography}
\end{document}